\documentclass[runningheads]{llncs}
\usepackage{graphicx}

\usepackage{tikz}
\usepackage{comment}
\usepackage{amsmath,amssymb} 
\usepackage{color}

\usepackage{graphicx}
\usepackage{amsmath}
\usepackage{amssymb}
\usepackage{booktabs}
\usepackage{tabularx}
\usepackage{multirow}
\usepackage{times}
\usepackage{epsfig}
\usepackage{subcaption}
\usepackage{dcolumn}

\usepackage[pagebackref,breaklinks,colorlinks]{hyperref}

\usepackage[capitalize]{cleveref}
\crefname{section}{Sec.}{Secs.}
\Crefname{section}{Section}{Sections}
\Crefname{table}{Table}{Tables}
\crefname{table}{Tab.}{Tabs.}
\crefname{figure}{Fig.}{Figs.}
\Crefname{figure}{Figure}{Figures}

\usepackage[accsupp]{axessibility}  

\definecolor{gold}{rgb}{0.83, 0.69, 0.22}
\definecolor{silver}{rgb}{0.75, 0.75, 0.75}
\definecolor{bronze}{rgb}{0.8, 0.5, 0.2}
\newcommand{\first}[1]{%
    {#1\raisebox{0.8pt}{\scriptsize \color{gold} \textcircled{\raisebox{-0.8pt}{1}}}}%
}
\newcommand{\second}[1]{%
    {#1\raisebox{0.8pt}{\scriptsize \color{silver} \textcircled{\raisebox{-0.8pt}{2}}}}%
}
\newcommand{\third}[1]{%
    {#1\raisebox{0.8pt}{\scriptsize \color{bronze} \textcircled{\raisebox{-0.8pt}{3}}}}%
}

\newcolumntype{d}[1]{D{.}{.}{#1}}

\begin{document}
\pagestyle{headings}
\mainmatter
\def\ECCVSubNumber{6990}  

\title{FEAR: Fast, Efficient, Accurate and Robust Visual Tracker} 

\titlerunning{ECCV-22 submission ID \ECCVSubNumber} 
\authorrunning{ECCV-22 submission ID \ECCVSubNumber} 
\author{Anonymous ECCV submission}
\institute{Paper ID \ECCVSubNumber}

\titlerunning{FEAR: Fast, Efficient, Accurate and Robust Visual Tracker}
%
\author{Vasyl Borsuk\inst{1,2 *} \and
Roman Vei\inst{1,2 *} \and
Orest Kupyn\inst{1,2} \and
Tetiana Martyniuk\inst{1,2} \and
Igor Krashenyi\inst{1,2} \and
Ji\v{r}i Matas\inst{3}}
\authorrunning{V. Borsuk et al.}
%
\institute{
Ukrainian Catholic University\\
\email{\{borsuk, vey, kupyn, t.martynyuk, igor.krashenyi\}@ucu.edu.ua}
\and
Piñata Farms, Los Angeles, USA\\
\email{\{vasyl, roman, orest, tetiana, igor\}@pinatafarm.com}
\and
Visual Recognition Group, Center for Machine Perception, FEE, CTU in Prague \\
\email{matas@cmp.felk.cvut.cz}
}
\maketitle

\def\thefootnote{*}\footnotetext{These authors contributed equally to this work.}

\begin{abstract}
We present FEAR, a family of \textbf{f}ast, \textbf{e}fficient, \textbf{a}ccurate, and \textbf{r}obust Siamese visual trackers. 
We present a novel and efficient way to benefit from dual-template representation for object model adaption, which incorporates temporal information with only a single learnable parameter.
We further improve the tracker architecture with a pixel-wise fusion block.
By plugging-in sophisticated backbones with the abovementioned modules, FEAR-M and FEAR-L trackers surpass most Siamese trackers on several academic benchmarks in both accuracy and efficiency.
Employed with the lightweight backbone, the optimized version FEAR-XS offers more than 10 times faster tracking than current Siamese trackers while maintaining near state-of-the-art results. 
FEAR-XS tracker is \textbf{2.4x} smaller and \textbf{4.3x} faster than LightTrack with superior accuracy. 
In addition, we expand the definition of the model efficiency by introducing FEAR benchmark that assesses energy consumption and execution speed. 
We show that energy consumption is a limiting factor for trackers on mobile devices.
Source code, pretrained models, and evaluation protocol are available at \url{https://github.com/PinataFarms/FEARTracker}.
\end{abstract}

\section{Introduction}

\begin{figure}[t]
\centering
\begin{minipage}[c]{0.45\textwidth}
  \includegraphics[width=0.9\linewidth]{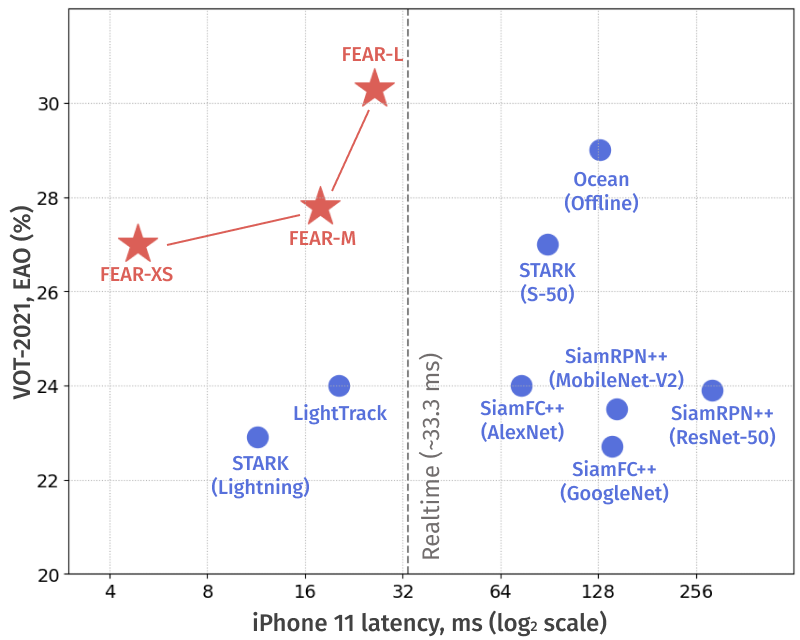} 
\end{minipage}
\begin{minipage}[c]{0.54\textwidth}
   \caption{\textbf{The EAO-Latency trade-off plot}. Compared to other state-of-the-art approaches (shown in blue), the FEAR-XS tracker (in red) achieves superior or comparable quality (EAO) while attaining outstanding speed on mobile devices; FEAR-L (in red) tracker runs in real-time on iPhone 11 and shows the best performance in EAO on VOT-2021.}
\end{minipage}  
 \label{fig:cover}
\end{figure}

Visual object tracking is a highly active research area of computer vision with many applications such as autonomous driving \cite{autonomous}, surveillance \cite{surveillance}, augmented reality \cite{AR}, and robotics \cite{robotics}. 
Building a general system for tracking an arbitrary object in the wild using only information about the location of the object in the first frame is non-trivial due to occlusions, deformations, lighting changes, background cluttering, reappearance, etc. \cite{OTB}.
Real-world scenarios often require models to be deployed on the edge devices with hardware and power limitations, adding further complexity. 
Thus, developing a robust tracking algorithm has remained a challenge.

The recent adoption of deep neural networks, specifically Siamese networks \cite{Siam2015}, has led to significant progress in visual object tracking \cite{SiamFC}, \cite{SiamRPN}, \cite{SiamFC++}, \cite{SiamRPN++}, \cite{DaSiamRPN}, \cite{SiamDW}, \cite{Ocean}. 
One of the main advantages of Siamese trackers is the possibility of end-to-end offline learning. 
In contrast, methods incorporating online learning \cite{ATOM}, \cite{DiMP}, \cite{MDNet} increase computational complexity to an unacceptable extent for real-world scenarios \cite{survey2}. 

Current state-of-the-art approaches for visual object tracking achieve high results on several benchmarks \cite{VOT}, \cite{VOT2020} at the cost of heavy computational load. 
Top-tier visual trackers like SiamRPN++ \cite{SiamRPN++} and Ocean \cite{Ocean} exploit complex feature extraction and cross-correlation modules, resulting in 54M parameters and 49 GFLOPs, and 26M parameters and 20 GFLOPs, respectively. 
Recently, STARK \cite{STARK} introduced a transformer-based encoder-decoder architecture for visual tracking with 23.3M parameters and 10.5 GFLOPs.
The large memory footprint cannot satisfy the strict performance requirements of real-world applications.
Employing a mobile-friendly backbone into the Siamese tracker architecture does not lead to a significant boost in the inference time, as most memory and time-consuming operations are in the decoder or bounding box prediction modules (see \Cref{t:flops}).
Therefore, designing a lightweight visual object tracking algorithm, efficient across a wide range of hardware, remains a challenging problem.
Moreover, it is essential to incorporate temporal information into the algorithm to make a tracker robust to pose, lighting, and other object appearance changes.  
This usually assumes adding either dedicated branches to the model \cite{STARK}, or online learning modules \cite{DiMP}. 
Either approach results in extra FLOPs that negatively impact the run-time performance.

We introduce a novel lightweight tracking framework, \emph{FEAR tracker}, that efficiently solves the above-mentioned problems.
We develop a single-parameter \emph{dual-template} module which allows to learn the change of the object appearance on the fly without any increase in model complexity, mitigating the memory bottleneck of recently proposed online learning modules \cite{ATOM}, \cite{DiMP}, \cite{KYS}, \cite{prDIMP}.
This module predicts the likelihood of the target object being close to the center of the search image, thus allowing to select candidates for the template image update.
Furthermore, we interpolate the online selected dynamic template image feature map with the feature map of the original static template image in a learnable way. 
This allows the model to capture object appearance changes during inference.
We optimize the neural network architecture to perform more than 10 times faster than most current Siamese trackers.
Additionally, we design an extra lightweight FEAR-XS network that achieves real-time performance on mobile devices while still surpassing or achieving comparable accuracy to the state-of-the-art deep learning methods.

The main contributions of the paper are:
\begin{itemize}

\item  A novel \textbf{dual-template representation} for object model adaptation.
The first template, static, anchors the original visual appearance
and thus prevents drift and, consequently, adaptation-induced failures.
The other is dynamic; its state reflects the current acquisition conditions and object appearance. 
Unlike STARK \cite{STARK}, which incorporates additional temporal information by introducing \textit{a separate score prediction head}, we introduce a \textit{parameter-free} similarity module as a template update rule, optimized with the rest of the network.
We show that a learned convex combination of the two templates is effective for tracking on multiple benchmarks.

    

\item  A lightweight tracker that combines a compact feature extraction network, the dual-template representation, and pixel-wise fusion blocks.
The resulting {\bf FEAR-XS tracker} runs at 205 FPS on iPhone 11, 4.2$\times$ faster than LightTrack \cite{LightTrack} and 26.6$\times$ faster than Ocean \cite{Ocean}, with high accuracy on multiple benchmarks - no state-of-the-art tracker is at the same time more accurate and faster than any of FEAR trackers. Besides, the algorithm is highly energy-efficient.

\item We introduce \textbf{FEAR benchmark} - a new tracker efficiency benchmark and protocol. Efficiency is defined in terms of both energy consumption and execution speed.
Such aspect of vision algorithms, important in real-world use, has not been benchmarked before. 
We show that current state-of-the-art trackers show high instantaneous speed when evaluated over a small test set, but slow down over time when processing large number of samples, as the device overheats when the tracker is not energy-efficient.
In that sense, FEAR family fills the gap between speed and accuracy for real-time trackers.
\end{itemize}




\section{Related Work}
\textbf{Visual Object Tracking.}
Conventional tracking benchmarks such as annual VOT challenges \cite{VOT} and the Online Tracking Benchmark \cite{OTB} have historically been dominated by hand-crafted features-based solutions \cite{Matas}, \cite{KCF}, \cite{Staple}.
With the rise of deep learning, they lost popularity constituting only 14\% of VOT-ST2020 \cite{VOT2020} participant models.
Lately, short-term visual object tracking task \cite{VOT2020} was mostly addressed using either discriminatory correlation filters \cite{ECO}, \cite{DiMP}, \cite{ATOM}, \cite{DCFST}, \cite{LADCF}, \cite{AFOD} or Siamese neural networks \cite{Ocean}, \cite{SiamRPN}, \cite{SiamRPN++}, \cite{DaSiamRPN}, \cite{SiamFC++}, \cite{SiamDW}, \cite{GOTURN}, as well as both combined \cite{RPT}, \cite{OceanPlus}, \cite{AlphaRef}. 
Moreover, success  of visual transformer networks for image classification \cite{ViT} has resulted in new high-scoring models \cite{STARK}, \cite{TransT}, \cite{TrMeetsTr} for tracking.

\textbf{Siamese trackers.} 
Trackers based on Siamese correlation networks perform tracking based on offline learning of a matching function. 
This function acts as a similarity metric between the features of the template image and the cropped region of the candidate search area.
Siamese trackers initially became popular due to their impressive trade-off between accuracy and efficiency \cite{SINT}, \cite{SiamFC}, \cite{RASNet}, \cite{SiamRPN}, \cite{DaSiamRPN}; however, they could not keep up with the accuracy of online learning methods \cite{ATOM}, \cite{DiMP}. 
With recent modeling improvements, Siamese-based trackers \cite{STARK}, \cite{Ocean} hold winning positions on the most popular benchmarks \cite{VOT}, \cite{GOT10k}, \cite{LaSOT}.

One of the state-of-the-art methods, Ocean \cite{Ocean}, incorporates FCOS \cite{FCOS} anchor-free object detection paradigm for tracking, directly regressing the distance from the point in the classification map to the corners of the bounding box.
Another state-of-the-art approach, STARK \cite{STARK}, introduces a transformer-based encoder-decoder in a Siamese fashion: flattened and concatenated search and template feature maps serve as an input to the transformer network. 

Neither of the forenamed state-of-the-art architectures explicitly addresses the task of fast, high-quality visual object tracking across the wide variety of GPU architectures.

Recently, LightTrack \cite{LightTrack} made a considerable step towards performant tracking on mobile, optimizing for FLOPs as well as model size via NAS \cite{nas}, \cite{detnas}. 
Still, FLOP count does not always reflect actual inference time \cite{FBNet}.

\textbf{Efficient Neural Networks.} Designing efficient and lightweight neural networks optimized for inference on mobile devices has attracted much attention in the past few years due to many practical applications. 
SqueezeNet \cite{SqueezeNet} was one of the first works focusing on reducing the size of the neural network.
They introduced an efficient downsampling strategy, extensive usage of 1x1 Convolutional blocks, and a few smaller modules to decrease the network size significantly. 
Furthermore, SqueezeNext \cite{SqueezeNext} and ShiftNet \cite{shift} achieve extra size reduction without any significant drop of accuracy. 
Recent works focus not only on the size but also on the speed, optimizing FLOP count directly. 
MobileNets introduce new architecture components: MobileNet \cite{MobileNet} uses depth-wise separable convolutions as a lightweight alternative to spatial convolutions, and MobileNet-v2 \cite{mobilenetv2} adds memory-efficient inverted residual layers.
ShuffleNet \cite{shufflenet} utilizes group convolutions and shuffle operations to reduce the FLOP count further. 
More recently, FBNet \cite{FBNet} also takes the hardware design into account, creating a family of mobile-optimized CNNs using neural architecture search. 

For FEAR trackers, we followed best practices for designing efficient and flexible neural network architecture. For an extremely lightweight version, where possible, we used depth-wise separable convolutions instead of regular ones and designed the network layers such that the Conv-BN-ReLU blocks could be fused at the export step.

\section{The method}

\begin{figure*}[t!]
  \includegraphics[width=0.99\textwidth]{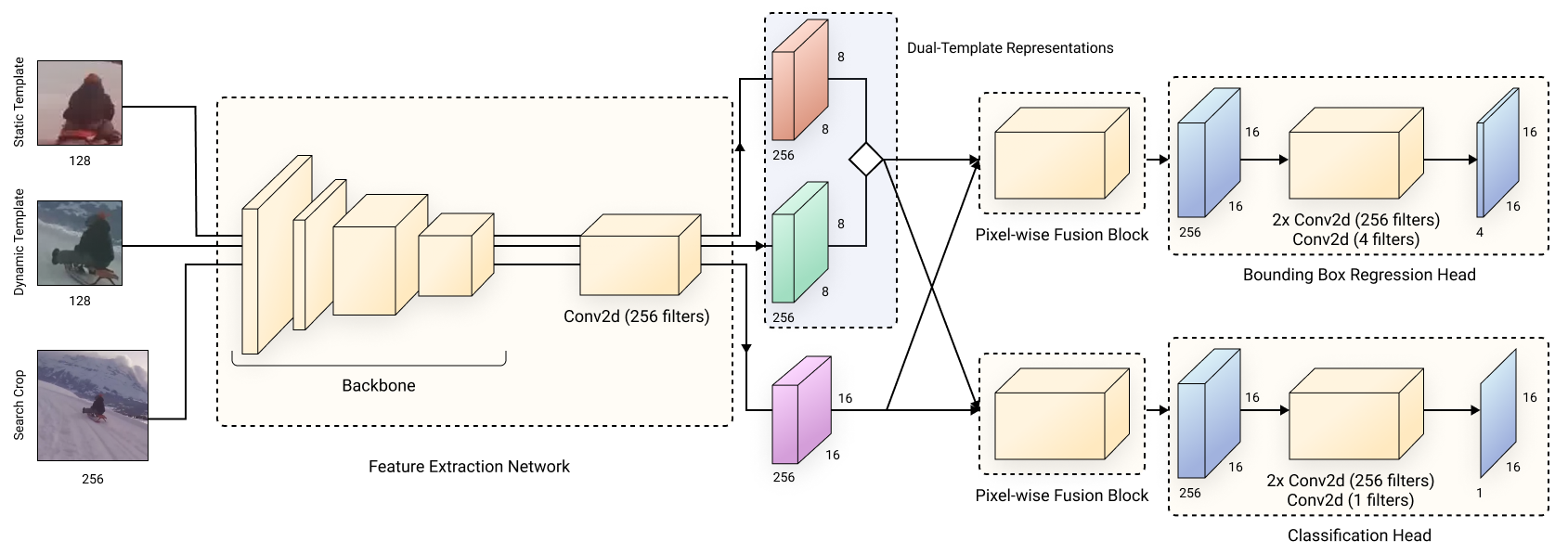}
  \caption{\textbf{The FEAR network architecture} consists of 5 components: feature extraction network, dual-template representation, pixel-wise fusion blocks, and bounding box and classification heads. 
  The CNN backbone extracts feature representations from the template and search images. 
  The dual-template representation allows for a single-parameter dynamic template update (see \cref{fig:update}).
  The pixel-wise fusion block effectively combines template and search image features (see \cref{fig:mobile_corr}). 
  The bounding box and classification heads make the final predictions for the box location and its presence, respectively.}
  \label{fig:arch}
\end{figure*}

FEAR tracker is a single, unified model composed of a feature extraction network, a dual-template representation, pixel-wise fusion blocks, and task-specific subnetworks for bounding box regression and classification. 
Given a static template image, $I_T$, a search image crop, $I_S$, and a dynamic template image, $I_{d}$, the feature extraction network yields the feature maps over these inputs.
The template feature representation is then computed as a linear interpolation between static and dynamic template image features. 
Next, it is fused with the search image features in the pixel-wise fusion blocks and passed to the classification and regression subnetworks. 
Every stage is described in detail further on, see the overview of the FEAR network architecture in \Cref{fig:arch}.

\subsection{FEAR Network Architecture}


\textbf{Feature Extraction Network.} Efficient tracking pipeline requires a flexible, lightweight, and accurate feature extraction network. 
Moreover, the outputs of such backbone network should have high enough spatial resolution to have optimal feature capability of object localization \cite{SiamRPN++} while not increasing the computations for the consecutive layers. 
Most of the current Siamese trackers \cite{Ocean}, \cite{SiamRPN++} increase the spatial resolution of the last feature map, which significantly degrades the performance of successive layers. 
We observe that keeping the original spatial resolution significantly reduces the computational cost of both backbone and prediction heads, as shown in \Cref{t:flops}.

\begin{table}[h]
\renewcommand{\arraystretch}{0.95}
\centering
\begin{tabular}{l||c|c}
Model architecture & Backbone & Prediction heads\\
 & GigaFLOPs & GigaFLOPs\\
\hline
FEAR-XS tracker & 0.318 & 0.160\\
FEAR-XS tracker $\uparrow$ & 0.840 & 0.746\\
OceanNet & 4.106 & 1.178 \\
OceanNet $\uparrow$ (original) & 14.137 & 11.843 \\
\end{tabular}
\vspace{0.75em}
\caption{GigaFLOPs, per frame, of the FEAR tracker and OceanNet \cite{Ocean} architectures; $\uparrow$ indicates the increased spatial resolutions of the backbone. We show in \Cref{section:ablations} that upscaling has a negligible effect on accuracy while increasing FLOPs significantly.} 
\label{t:flops}
\end{table}

We use the first four stages of the neural network pretrained on the ImageNet \cite{ImageNet} as a feature extraction module. 
The FEAR-M tracker adopts the vanilla ResNet-50 \cite{ResNet} as a backbone, and the FEAR-L tracker incorporates the RegNet \cite{xu2021regnet} backbone to pursue the state-of-the-art tracking quality, yet remaining efficient.

The output of the backbone network is a feature map of stride \textit{16} for the template and search images. 
To map the depth of the output feature map to a constant number of channels, we use a simple \textit{AdjustLayer} which is a combination of Convolutional and Batch Normalization \cite{BatchNorm} layers. 

To shift towards being more efficient during inference on mobile devices, for the mobile version of our tracker - FEAR-XS - we utilize the FBNet \cite{FBNet} family of models designed via NAS. 
\Cref{t:flops} and \Cref{fig:battery-online} demonstrate that even a lightweight encoder does not improve the model efficiency of modern trackers due to the complex prediction heads. Thus, designing a lightweight and accurate decoder is still a challenge.

\begin{figure}[t]
  \centering
  \includegraphics[width=0.7\textwidth]{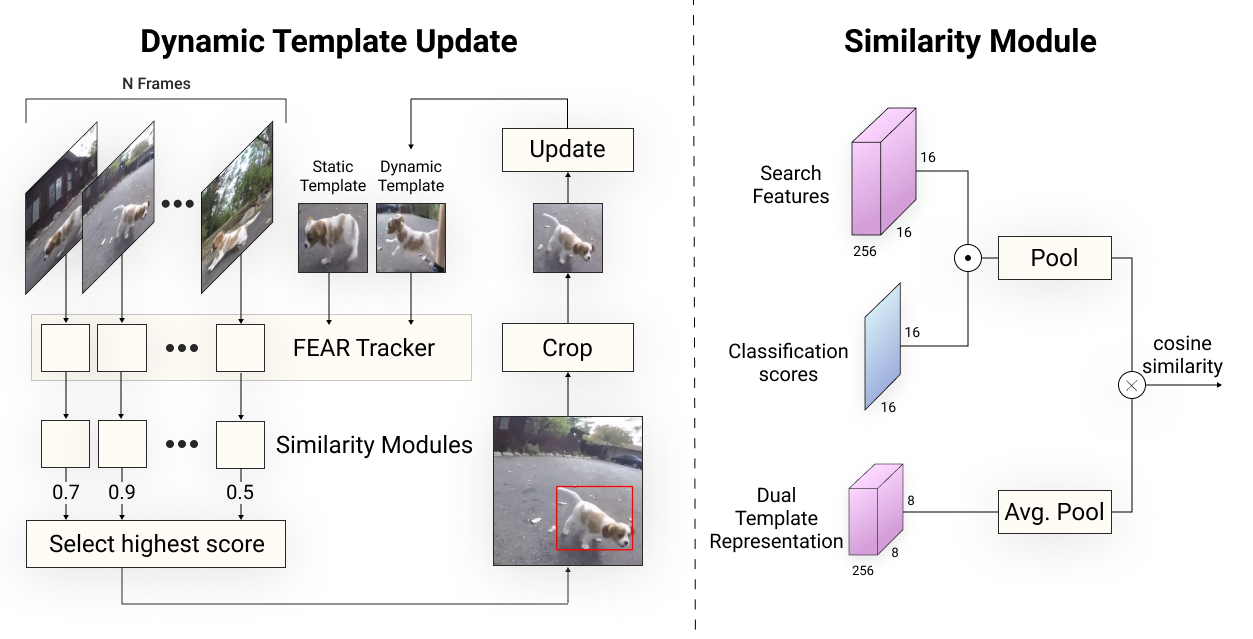}
  \caption{\textbf{Dynamic Template update.} We compare the average-pooled dual-template representation with the search image embedding using cosine similarity, and dynamically update the template representation when object appearance changes dramatically.}
  \label{fig:update}
\end{figure}

\textbf{The dual-template representation} with a dynamic template update allows the model to capture the appearance changes of objects during inference without the need to perform optimization on the fly. 
The general scheme of the Dynamic Template Update algorithm is shown in \Cref{fig:update}.
In addition to the main static template $I_T$ and search image $I_S$, we randomly sample a dynamic template image, $I_{d}$, from a video sequence during model training to capture the object under various appearances. 
We pass $I_{d}$ through the feature extraction network, and the resulting feature map, $F_{d}$, is linearly interpolated with the main template feature map $F_T$ via a learnable parameter $w$:
\begin{equation} \label{eq1}
F_T' = (1 - w)F_T + w F_{d}
\end{equation}

We further pass $F_T'$ and $F_S$ to the Similarity Module that computes cosine similarity between the dual-template and search image embeddings. 
The search image embedding $e_S$ is obtained via the Weighted Average Pooling (WAP) \cite{WAP} of $F_S$ by the classification confidence scores; the dual-template embedding $e_T$ is computed as an Average Pooling \cite{AvgPool} of $F_T'$.

During inference, for every $N$ frames we choose the search image with the highest cosine similarity with the dual-template representation, and update the dynamic template with the predicted bounding box at this frame. 
In addition, for every training pair we sample a negative crop $I_N$ from a frame that does not contain the target object. 
We pass it through the feature extraction network, and extract the negative crop embedding $e_N$ similarly to the search image, via WAP.
We then compute Triplet Loss \cite{TripletLoss} with the embeddings $e_T, e_S, e_N$ extracted from $F_T'$, $F_S$ and $F_N$, respectively. 
This training scheme does provide a signal for the dynamic template scoring while also biasing the model to prefer more general representations.

Unlike STARK \cite{STARK}, which incorporates additional temporal information by introducing \textit{a separate score prediction head} to determine whether to update the dynamic template, we present a \textit{parameter-free} similarity module as a template update rule, optimized with the rest of the network.
Moreover, STARK concatenates the dynamic and static template features, increasing the size of a tensor passed to the encoder-decoder transformer resulting in more computations.
Our dual-template representation interpolates between the static and dynamic template features with \textit{a single learnable parameter}, not increasing the template tensor size.

In \Cref{section4}, we demonstrate the efficiency of our method on a large variety of academic benchmarks and challenging cases. 
The dual-template representation module allows the model to efficiently encode the temporal information as well as the object appearance and scale change. The increase of model parameters and FLOPs is small and even negligible, making it almost a cost-free temporal module.

\begin{figure}[t]
  \centering
  \includegraphics[width=0.6\textwidth]{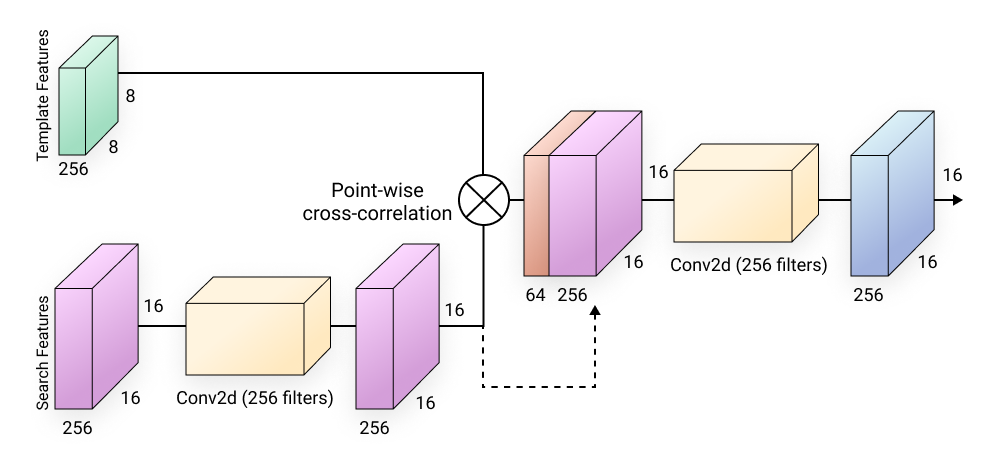}
  \caption{\textbf{The pixel-wise fusion block}. The search and template features are combined using the point-wise cross-correlation module and enriched with search features via concatenation. The output is then forwarded to regression heads. }
  \label{fig:mobile_corr}
\end{figure}

\textbf{Pixel-wise fusion block.}\label{para} The cross-correlation module creates a joint representation of the template and search image features. 
Most existing Siamese trackers use either simple cross-correlation operation \cite{SiamFC}, \cite{SiamFC++}, \cite{SiamRPN} or more lightweight depth-wise cross-correlation \cite{SiamRPN++}. 
Recently, Alpha-Refine \cite{AlphaRef} avoided correlation window blurring effect by adopting the pixel-wise correlation as it ensures that each correlation map encodes information of a local region of the target.
Extending this idea, we introduce a pixel-wise fusion block which enhances the similarity information obtained via pixel-wise correlation with position and appearance information extracted from the search image (see \Cref{t:ablation}). 

We pass the search image feature map through a 3x3 Conv-BN-ReLU block, and calculate the point-wise cross-correlation between these features and template image features.
Then, we concatenate the computed correlation feature map with the search image features, and pass the result through a 1x1 Conv-BN-ReLU block to aggregate them. 
With this approach, learned features are more discriminative and can efficiently encode object position and appearance: see \cref{section:ablations} and \cref{t:ablation} for the detailed ablation study. 
The overall architecture of a pixel-wise fusion block is visualized in \cref{fig:mobile_corr}.


\textbf{Classification and Bounding Box Regression Heads.} The core idea of a bounding box regression head is to estimate the distance from each pixel within the target object's bounding box to the ground truth bounding box sides \cite{Ocean}, \cite{FCOS}.
Such bounding box regression takes into account all of the pixels in the ground truth box during training, so it can accurately predict the magnitude of target objects even when only a tiny portion of the scene is designated as foreground.
The bounding box regression network is a stack of two simple 3x3 Conv-BN-ReLU blocks. We use just two such blocks instead of four proposed in Ocean \cite{Ocean} to reduce computational complexity.
The classification head employs the same structure as a bounding box regression head. The only difference is that we use one filter instead of four in the last Convolutional block. This head predicts a 16x16 score map, where each pixel represents a confidence score of object appearance in the corresponding region of the search crop.

\subsection{Overall Loss Function:} Training a Siamese tracking model requires a multi-component objective function to simultaneously optimize classification and regression tasks. 
As shown in previous approaches \cite{Ocean}, \cite{STARK}, IoU loss \cite{IoUloss} and classification loss are used to efficiently train the regression and classification networks jointly. In addition, to train FEAR trackers, we supplement those training objectives with \textbf{triplet loss}, which enables performing Dynamic Template Update. As seen in the ablation study, it improves the tracking quality by $0.6\%$ EAO with only a single additional trainable parameter and marginal inference cost (see \Cref{t:ablation}). To our knowledge, this is a novel approach in training object trackers.

The triplet loss term is computed from template ($e_T$), search ($e_S$), and negative crop ($e_N$) feature maps:
 \begin{equation} \label{eq2}
L_t = \max\left \{ d(e_T, e_S) -d(e_T, e_N) + \textnormal{margin}, 0)) \right \},
\end{equation}
where $d(x_i, y_i) = \left \| x_i-y_i \right \|_2$.
The regression loss term is computed as:
 \begin{equation} \label{eq4}
L_{reg} = 1 -\sum_i IoU(t_{reg}, p_{reg}),
\end{equation}
where $t_{reg}$ denotes the target bounding box, $p_{reg}$ denotes the predicted bounding box, and $i$ indexes the training samples.
For classification loss term, we use Focal Loss \cite{FocalLoss}:
 \begin{equation} \label{eq5}
L_c = -(1 - p_t)^\gamma \mathrm{log}(p_t),
\quad
p_t =
    \begin{cases}
      p & \text{if y = 1,}\\
      1 - p & \text{otherwise.}
    \end{cases}
\end{equation}

In the above, $y \in \{-1; 1\}$ is a GT class, and $0 \le p \le 1$ is the predicted probability for the class $y=1$.
The overall loss function is a linear combination of the three components:
 \begin{equation} \label{eq6}
L = \lambda_1 * L_t + \lambda_2 * L_{reg} + \lambda_3 * L_c.
\end{equation}
In practice, we use 0.5, 1.0, 1.0 as $\lambda_1$, $\lambda_2$, $\lambda_3$, respectively. 

\section{Experimental Evaluation}\label{section4}

\subsection{Implementation Details}

\textbf{Training.} We implemented all of models using PyTorch \cite{pytorch}. The backbone network is initialized using the pretrained weights on ImageNet. All the models are trained using \textbf{4~RTX~A6000} GPUs, with a total batch size of 512. We use ADAM \cite{ADAM} optimizer with a learning rate = $4 * 10^{-4}$ and a plateau learning rate reducer with a factor = 0.5 every 10 epochs monitoring the target metric (mean IoU). Each epoch contains $10^6$ image pairs. The training takes 5 days to converge.
For each epoch, we randomly sample 20,000 images from LaSOT \cite{LaSOT}, 120,000 from COCO \cite{COCO}, 400,000 from YoutubeBB \cite{YouTubeBB}, 320,000 from GOT10k \cite{GOT10k} and 310,000 images from the ImageNet dataset \cite{ImageNet}, so, overall, 1,170,000 images are used in each epoch. 

From each video sequence in a dataset, we randomly sample a template frame $I_T$ and search frame $I_S$ such that the distance between them is $d = 70$ frames. 
Starting from the 15\textit{th} epoch, we increase $d$ by 2 every epoch. 
It allows the network to learn the correlation between objects on easier samples initially and gradually increase complexity as the training proceeds.
A dynamic template image is sampled from the video sequence between the static template frame and search image frame. 
For the negative crop, where possible, we sample it from the same frame as the dynamic template but without overlap with this template crop; otherwise, we sample the negative crop from another video sequence.
The value for $d$ was found empirically. 
It is consistent with the note in TrackingNet \cite{trackingnet} that any tracker is reliable within 1 second. 
Our observations are that the appearance of objects does not change dramatically over 2 seconds (60 frames), and we set $d = 70$ as a trade-off between the inference speed and the amount of additionally incorporated temporal information.

\textbf{Preprocessing.} We extract template image crops with an additional offset of $20\%$ around the bounding box. 
Then, we apply a light shift (up to 8px) and random scale change (up to $5\%$ on both sides) augmentations, pad image to the square size with the mean RGB value of the crop, and resize it to the size of 128x128 pixels.
We apply the same augmentations with a more severe shift (up to 48px) and scale (between $65\%$ and $135\%$ from the original image size) for the search and negative images. 
Next, the search image is resized to 256x256 pixels with the same padding strategy as in the template image.

Finally, we apply random photometric augmentations for both search and template images to increase model generalization and robustness under different lighting and color conditions \cite{albumentations}.

\textbf{Testing:} During inference, tracking follows the same protocols as in \cite{SiamFC}, \cite{SiamRPN}. 
The static template features of the target object are computed once at the first frame. 
The dynamic template features are updated every 70 frames and interpolated with the static template features. 
These features are combined with the search image features in the correlation modules, regression, and classification heads to produce the final output.

\subsection{Tracker efficiency benchmark} 

\begin{figure}[t!]
\centering
\includegraphics[width=0.6\textwidth]{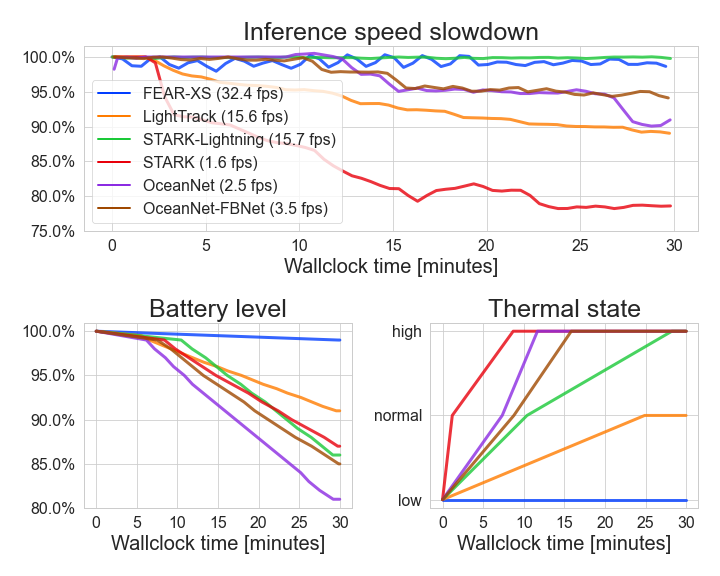}
\caption{\textbf{Online Efficiency Benchmark on iPhone 8:} battery consumption, device thermal state, and inference speed degradation over time. FEAR-XS tracker does not change the thermal state of the device and has a negligible impact on the battery level. Transformer-based trackers have a battery level drop comparable to the Siamese trackers, reaching a high thermal state in less than 10 minutes of online processing.}
\label{fig:battery-online}
\end{figure}

\paragraph{Setup:} Mobile devices have a limited amount of both computing power and energy available to execute a program. Most current benchmarks measure only runtime speed without taking into account the energy efficiency of the algorithm, which is equally important in a real-world scenario. Thus, we introduce the \textbf{FEAR Benchmark} to estimate the effect of tracking algorithms on mobile device battery and thermal state and its impact on the processing speed over time. It measures the energy efficiency of trackers with online and offline evaluation protocols - the former to estimate the energy consumption for the real-time input stream processing and the latter to measure the processing speed of a constant amount of inputs.

The online evaluation collects energy consumption data by simulating a real-time (30 FPS) camera input to the neural network for 30 minutes.
The tracker cannot process more frames than the specified FPS even if its inference speed is faster, and it skips inputs that cannot be processed on-time due to the slower processing speed. 
We collect battery level, device's thermal state, and inference speed throughout the whole experiment. 
The thermal state is defined by Apple in the official Thermal state iOS API \cite{thermalstate}. 
The \textit{high} thermal state refers to a critical thermal state when system's performance is significantly reduced to cool it down.
The performance loss due to heat causes trackers to slow down, making it a critical performance metric when deployed to mobile devices.
FEAR benchmark takes care of these issues providing fair comparison (see \cref{fig:battery-online}).

The offline protocol measures the inference speed of trackers by simulating a constant number of random inputs for the processing.
All frames are processed one by one without any inference time restrictions.
Additionally, we perform a model warmup before the experiment, as the first model executions are usually slower. We set the number of warmup iterations and inputs for the processing to 20 and 100, respectively.

In this work, we evaluate trackers on iPhone 7, iPhone 8 Plus, iPhone 11, and Pixel 4. All devices are fully charged before the experiment, no background tasks are running, and the display is set to the lowest brightness to reduce the energy consumption of hardware that is not involved in computations.

We observe that algorithms that reach the high system thermal state get a significant drop in the processing speed due to the smaller amount of processing units available. The results prove that the tracking speed is dependent on the energy efficiency, and both should be taken into account. 

\textbf{Online efficiency benchmark:} \cref{fig:battery-online} summarizes the online benchmark results on iPhone 8. The upper part of the plot demonstrates the degradation of inference speed over time. We observe that FEAR-XS tracker and STARK-Lightning \cite{stark_lightning} backbone do not change inference speed over time, while LightTrack \cite{LightTrack} and OceanNet \cite{Ocean} start to process inputs slower. Also, transformer network STARK-S50 degrades significantly and becomes 20\% slower after 30 minutes of runtime. The lower part of the figure demonstrates energy efficiency of FEAR-XS tracker against competitors and its negligible impact on device thermal state. STARK-S50 and Ocean overheat device after 10 minutes of execution, LightTrack slightly elevates temperature after 24 minutes, STARK-Lightning overheats device after 27 minutes, while FEAR-XS tracker keeps device in a low temperature. Moreover, Ocean with a lightweight backbone FBNet \cite{FBNet} still consumes lots of energy and produces heat due to complex and inefficient decoder.

Additionally, we observe that STARK-Lightning reaches high thermal state without performance drop. Modern devices have a special hardware, called Neural Processing Unit (NPU), designed specifically for neural network inference. The Apple Neural Engine (ANE) is a type of NPU that accelerates neural network operations such as convolutions and matrix multiplies. STARK-Lightning is a transformer based on simple matrix multiplications that are efficiently computed by ANE and thus do not slow down over time.

\begin{figure}[t!]
\centering
  \includegraphics[width=0.55\textwidth]{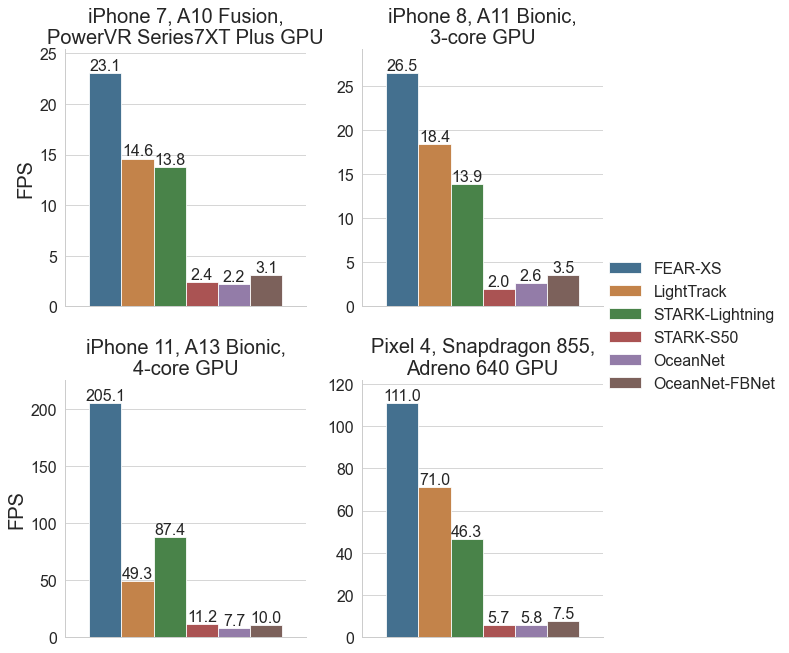}
  \caption{\textbf{Offline Efficiency Benchmark:} mean FPS on a range of mobile GPU architectures. FEAR-XS tracker has superior processing speed on all devices while being an order of magnitude faster on a modern GPU -- Apple A13 Bionic. 
}
  \label{fig:battery-offline}
\end{figure}

\textbf{Offline efficiency benchmark:} We summarize the results of offline benchmark in \Cref{fig:battery-offline}. 
We observe that FEAR-XS tracker achieves ~1.6 times higher FPS than LightTrack \cite{LightTrack} on iPhone 7 (A10 Fusion and PowerVR Series7XT GPU), iPhone 8 (A11 Bionic with 3-core GPU) and Google Pixel 4 (Snapdragon 855 and Adreno 640 GPU). 
Furthermore, FEAR-XS tracker is more than 4 times faster than LightTrack on iPhone 11 (A11 Bionic with 4-core GPU). 
FEAR-XS tracker achieves more than ~10 times faster inference than OceanNet \cite{Ocean} and STARK \cite{STARK} on all aforementioned mobile devices. 
Such low inference time makes FEAR-XS tracker a very cost-efficient candidate for use in resource-constrained applications.



\subsection{Comparison with the state-of-the-art}

\begin{table}[tb]
\small
\renewcommand{\arraystretch}{0.95}
\centering
\begin{tabular}{c|l|l|l}
\noalign{\smallskip}
\quad FPS \quad \quad & Success Score & Precision Score & Success Rate\\
\hline
30 & 0.618 & 0.753 & 0.780\\
240 & \textbf{0.655} & \textbf{0.816} & \textbf{0.835}\\
\noalign{\smallskip}
\end{tabular}
\caption{ \textbf{Extremely High FPS Tracking Matters}. The metrics were computed from the same set of frames on 30 and 240 fps NFS benchmark\cite{nfs}. FEAR-XS, tracking in over 200 fps, achieves superior performance than trackers limited to 30 fps by incorporating additional temporal information from intermediate frames.
}
\label{t:fps}
\end{table}

\begin{table*}[tb]
\footnotesize
\caption{\textbf{Comparison of {FEAR} and the state-of-the-art trackers} on common benchmarks: VOT-ST2021\cite{VOT},  GOT-10K\cite{GOT10k},
LaSOT\cite{LaSOT}, and NFS\cite{nfs}. 
FEAR trackers use much fewer parameters, achieves higher FPS; their accuracy and robustness is on par with the best. 
\first{}, \second{} and \third{} indicate the top-3 trackers
}\label{tab:sota}
\setlength{\tabcolsep}{3pt}
	\begin{subtable}[t]{.99\linewidth}\centering
		{\scriptsize
\resizebox{\textwidth}{!}{\begin{tabular}{r|ccccccccc|ccc}
\noalign{\smallskip}
  & SiamFC++ & SiamRPN++ & SiamRPN++ & ATOM & KYS & Ocean & STARK & STARK & LightTrack & \textbf{FEAR-XS} & \textbf{FEAR-M} & \textbf{FEAR-L}\\
 & (GoogleNet) & (MobileNet-V2) & (ResNet-50) & & & (offline) & (S50) & (lightning) & & & &\\
 & \cite{SiamFC++} & \cite{SiamRPN++} & \cite{SiamRPN++} & \cite{ATOM} & \cite{KYS} & \cite{Ocean} & \cite{STARK} & \cite{stark_lightning} & \cite{LightTrack} & \\
\hline
EAO ↑ & 0.227 & 0.235 & 0.239 & 0.258 & 0.274 & \second{0.290} & 0.270 & 0.226  & 0.240 & 0.270 & \third{0.278} & \first{0.303}\\
Accuracy ↑ & 0.418 & 0.432 & 0.438 & 0.457 & 0.453 & \second{0.479} & 0.464 & 0.433  & 0.417 & 0.471 & \third{0.476} & \first{0.501}\\
Robustness ↑ & 0.667 & 0.656 & 0.668 & 0.691 & \second{0.736} & \third{0.732} & 0.719 & 0.627 & 0.684 & 0.708 & 0.728 & \first{0.755}\\
\midrule
iPhone 11 FPS ↑ & 7.11 & 6.86 & 3.49 & - & - & 7.72 & 11.2 & \second{87.41} & 49.13 & \first{205.12} & \third{56.20} & 38.3\\
Parameters (M) ↓ & 12.71 & 11.15 & 53.95 & - & - & 25.87 & 23.34 & \third{2.28} & \second{1.97} & \first{1.37} & 9.67 & 33.65\\
{Memory (MB) ↓} & 24.77 & 21.63 & 103.74 & - & - & 102.81 & 109.63 & \third{6.28} & \second{4.11} & \first{3.00} & 18.82 & 66.24\\
{Peak memory (MB) ↓} & 34.17 & 31.39 & 192.81 & - & - & 119.51 & 295.97 & 30.69 & \first{9.21} & \second{10.10} & \third{25.88} & 85.97\\
\end{tabular}}
\caption{ \textbf{VOT-ST2021} \cite{VOT}}
\label{t:vot_st_2021_benchmark}

}	
	\end{subtable}
	\hfill
	\begin{subtable}[t]{.99\linewidth}\centering
		{
\centering
\resizebox{\textwidth}{!}{\begin{tabular}{r|cccccc|ccc}
 & SiamRPN++ & ATOM & KYS & Ocean & STARK & LightTrack  & \textbf{FEAR-XS} & \textbf{FEAR-M} & \textbf{FEAR-L}\\
 & (ResNet-50) & & & (offline) & (S50) & & &\\
 & \cite{SiamRPN++} & \cite{ATOM} & \cite{KYS} & \cite{Ocean} & \cite{STARK} & \cite{LightTrack} & & &\\
\hline
Average Overlap ↑ & 0.518 & 0.556 & \third{0.636} & 0.592 & \first{0.672} & 0.611 & 0.619 & 0.623 & \second{0.645} \\
Success Rate ↑ & 0.618 & 0.634 & \second{0.751} & 0.695 & \first{0.761} & 0.710 & 0.722 & 0.730 & \third{0.746} \\
\end{tabular}}
\caption{ \textbf{GOT-10K} \cite{GOT10k}}
\label{t:got10k_test_benchmark}


}	
	\end{subtable}
	\\
	\begin{subtable}[t]{.99\linewidth}\centering
	    {
\centering
\resizebox{\textwidth}{!}{\begin{tabular}{r|ccccccc|ccc}
 & SiamRPN++ & ATOM & KYS & Ocean & STARK & STARK & LightTrack & \textbf{FEAR-XS} & \textbf{FEAR-M} & \textbf{FEAR-L}\\
 & (ResNet-50) & & & (offline) & (S50) & (lightning) & & & \\
 & \cite{SiamRPN++} & \cite{ATOM} & \cite{KYS} & \cite{Ocean} & \cite{STARK} & \cite{stark_lightning} & \cite{LightTrack} & & &  \\
\hline
Success Score ↑ & 0.503& 0.491 & 0.541 & 0.505 & \first{0.668} & \second{0.586} & 0.523 & 0.535 & 0.546 & \third{0.579}\\
Precision Score ↑ & 0.496 & 0.483 & 0.539 & 0.517 & \first{0.701} & \third{0.579} & 0.515 & 0.545 & 0.556 & \second{0.609}\\
Success Rate ↑ & 0.593 & 0.566 & 0.640 & 0.594 & \first{0.778} & \second{0.690} & 0.596 & 0.641 & 0.638 & \third{0.686}\\
\end{tabular}}
\caption{ \textbf{LaSOT} \cite{LaSOT}}
\label{t:lasot_benchmark}


}	
	\end{subtable}
	\hfill
	\begin{subtable}[t]{.99\linewidth}\centering
		{
\resizebox{\textwidth}{!}{\begin{tabular}{r|ccccccc|ccc}
\noalign{\smallskip}
 & SiamRPN++ & ATOM & KYS & Ocean & STARK & STARK & LightTrack & \textbf{FEAR-XS} & \textbf{FEAR-M} & \textbf{FEAR-L}\\
 & (ResNet-50) & & & (offline) & (S50) & (lightning) & & & \\
 & \cite{SiamRPN++} & \cite{ATOM} & \cite{KYS} & \cite{Ocean} & \cite{STARK} & \cite{stark_lightning} & \cite{LightTrack} & & \\
\hline
Success Score ↑ & 0.596 & 0.592 & \third{0.634} & 0.573 & \first{0.681} & 0.628 & 0.591 & 0.614 & 0.622 & \second{0.658} \\
Precision Score ↑ & 0.720 & 0.711 & 0.766 & 0.706 & \first{0.825} & 0.754 & 0.730 & \third{0.768} & 0.745 & \second{0.814} \\
Success Rate ↑ & 0.748 & 0.737 & 0.795 & 0.728 & \first{0.860} & \third{0.796} & 0.743 & 0.788 & 0.788 & \second{0.834} \\
\end{tabular}}
\caption{\textbf{NFS} \cite{nfs}}
\label{t:nfs_benchmark}


}	
	\end{subtable}
\vspace{-2.0em}
\end{table*}

We compare FEAR trackers to existing state-of-the-art Siamese\cite{Ocean}, \cite{LightTrack}, \cite{SiamFC++}, \cite{SiamRPN++} and DCF\cite{ATOM}, \cite{DiMP}, \cite{KYS} trackers in terms of model accuracy, robustness and speed. We evaluate performance on two short-term tracking benchmarks: VOT-ST2021\cite{VOT}, GOT-10k\cite{GOT10k} and two long-term tracking benchmarks: LaSOT\cite{LaSOT}, NFS\cite{nfs}.
We provide three version of FEAR tracker: \textbf{FEAR-XS}, \textbf{FEAR-M} and \textbf{FEAR-XL}. The first one is a lightweight network optimized for on-device inference while two latter networks are more heavy and provide more accurate results.

\textbf{VOT-ST2021 Benchmark:} 
This benchmark consists of 60 short video sequences with challenging scenarios: similar objects, partial occlusions, scale and appearance change to address short-term, causal, model-free trackers. 
\Cref{t:vot_st_2021_benchmark} reports results on VOT-ST2021. It takes both Accuracy (A) and Robustness (R) into account to compute the \textit{bounding box} Expected Average Overlap metric (EAO) \cite{VOT} which is used to evaluate the overall performance.
FEAR-L tracker demonstrates 1.3\% higher EAO than Ocean \cite{Ocean} and outperforms trackers with online update, such as ATOM \cite{ATOM} and KYS \cite{KYS}, by 3\% EAO.
FEAR-XS tracker shows near state-of-the-art performance, outperforming LightTrack \cite{LightTrack} and STARK-Lightning \cite{stark_lightning} by 3\% and 4.4\% EAO, respectively, while having higher FPS. Also, it is only 2\% behind Ocean, yet having more than \textbf{18 times fewer parameters} than Ocean tracker and being \textbf{26 times faster} at model inference time (iPhone 11).

\Cref{t:vot_st_2021_benchmark} additionally reports model weights memory consumption and peak memory consumption during the forward pass in megabytes. LightTrack and STARK-Lightning model sizes are 4.11MB and 6.28MB, respectively, while FEAR-XS consumes only 3MB. During the forward pass, the peak memory usage of FEAR-XS is 10.1MB, LightTrack consumes slightly less (9.21MB) by using fewer filters in bounding box regression convolutional layers, and STARK-Lightning has 30.69MB peak memory usage due to memory-consuming self-attention blocks.

\textbf{GOT-10K Benchmark:} GOT-10K \cite{GOT10k} is a benchmark covering a wide range of different objects, their deformations, and occlusions. We evaluate our solution using the official GOT-10K submission page. FEAR-XS tracker achieves better results than LightTrack \cite{LightTrack} and Ocean \cite{Ocean}, while using 1.4 and 19 times fewer parameters, respectively. More details in the \Cref{t:got10k_test_benchmark}.

\textbf{LaSOT Benchmark:} LaSOT \cite{LaSOT} contains 280 video segments for long-range tracking evaluation. 
Each sequence is longer than 80 seconds in average making in the largest densely annotated long-term tracking benchmark. 
We report the \textit{Success Score} as well as \textit{Precision Score} and \textit{Success Rate}. As presented in \Cref{t:lasot_benchmark}, the \textit{Precision Score} of FEAR-XS tracker is 3\% and 2.8\% superior than LightTrack \cite{LightTrack} and Ocean \cite{Ocean}, respectively. Besides, the larger FEAR-M and FEAR-L trackers further improve \textit{Success Score} outperforming KYS \cite{KYS} by 0.5\% and 3.8\%.

\textbf{NFS Benchmark:} NFS \cite{nfs} dataset is a long-range benchmark, which has 100 videos (380K frames) captured with now commonly available higher frame rate (240 FPS) cameras from real world scenarios. 
\Cref{t:nfs_benchmark} presents that FEAR-XS tracker achieves better \textit{Success Score} (61.4\%), being 2.3\% and 4.1\% higher than LightTrack \cite{LightTrack} and Ocean \cite{Ocean}, respectively. Besides, FEAR-L tracker outperforms KYS \cite{KYS} by 2.4\% \textit{Success Score} and 4.8\% \textit{Precision Score}. Additionally, \Cref{t:fps} reports the impact of extremely high FPS video processing on accuracy, implying the importance of developing a fast tracker capable to process videos in higher FPS.


\subsection{Ablation Study}\label{section:ablations}

To verify the efficiency of the proposed method, we evaluate the effects of its different components on the VOT-ST2021 \cite{VOT} benchmark, as presented in \Cref{t:ablation}. 
The baseline model (\#1) consists of the FBNet backbone with an increased spatial resolution of the final stage, followed by a plain pixel-wise cross-correlation operation and bounding box prediction network. 
The performance of the baseline is 0.236 EAO and 0.672 Robustness. In \#2, we set the spatial resolution of the last stage to its original value and observe a negligible degradation of EAO while significantly increasing FPS on mobile. 
Adding our pixel-wise fusion blocks (\#3) brings a 3\% EAO improvement. 
This indicates that combining search image features and correlation feature maps enhances feature representability and improves tracking accuracy. 
Furthermore, the proposed dynamic template update module (\#4) also brings an improvement of 0.6\% in terms of EAO and 2.5\% Robustness, showing the effectiveness of this module. The pixel-wise fusion block and dynamic template update brought a significant accuracy improvements while keeping almost the same inference speed.
Note that the EAO metrics is calculated w.r.t. \textit{bounding box} tracking.

\begin{table}[t!]
\small
\renewcommand{\arraystretch}{0.95}
\centering
\begin{tabular}{r|l|c|c|c}
\noalign{\smallskip}
\# & Component & \quad EAO$\uparrow$ \quad & Robustness$\uparrow$ & iPhone 11 FPS$\uparrow$\\
\hline
1 & \textit{baseline} & 0.236  & 0.672 & 122.19\\
2 & + lower spatial resolution & 0.234 & 0.668 & 208.41\\
3 & + pixel-wise fusion block & 0.264 & 0.683 & 207.72\\
4 & + dynamic template update & 0.270 & 0.708 & 205.12 \\
\noalign{\smallskip}
\end{tabular}
\caption{ \textbf{FEAR-XS tracker -- Ablation study on VOT-ST2021 \cite{VOT}}.
}
\label{t:ablation}
\end{table}


\section{Conclusions} 
In this paper, we introduce the \emph{FEAR tracker} family - an efficient and powerful new Siamese tracking framework that benefits from novel architectural blocks.
We validate FEAR trackers performance on several popular academic benchmarks and show that the models near or exceed existing solutions while reducing the computational cost of inference. 
We demonstrate that the FEAR-XS model attains real-time performance on embedded devices with high energy efficiency.
Additionally, we introduce a novel tracker efficiency benchmark, where FEAR trackers demonstrate their energy efficiency and high inference speed, being more efficient and accurate than current state-of-the-art approaches at the same time.



\textbf{Acknowledgements.} We thank the Armed Forces of Ukraine for providing security to complete this work.


\clearpage
%
%
\bibliographystyle{splncs04}
\bibliography{egbib}

\begin{thebibliography}{10}
\providecommand{\url}[1]{\texttt{#1}}
\providecommand{\urlprefix}{URL }
\providecommand{\doi}[1]{https://doi.org/#1}

\bibitem{tensorflow2015-whitepaper}
Abadi, M., Agarwal, A., Barham, P., Brevdo, E., Chen, Z., Citro, C., Corrado,
  G.S., Davis, A., Dean, J., Devin, M., et~al.: Tensorflow: Large-scale machine
  learning on heterogeneous distributed systems. arXiv preprint
  arXiv:1603.04467  (2016)

\bibitem{Staple}
Bertinetto, L., Valmadre, J., Golodetz, S., Miksik, O., Torr, P.H.: Staple:
  Complementary learners for real-time tracking. In: Proceedings of the IEEE
  conference on computer vision and pattern recognition. pp. 1401--1409 (2016)

\bibitem{SiamFC}
Bertinetto, L., Valmadre, J., Henriques, J.F., Vedaldi, A., Torr, P.H.:
  Fully-convolutional siamese networks for object tracking. In: European
  conference on computer vision. pp. 850--865. Springer (2016)

\bibitem{DiMP}
Bhat, G., Danelljan, M., Gool, L.V., Timofte, R.: Learning discriminative model
  prediction for tracking. In: Proceedings of the IEEE/CVF International
  Conference on Computer Vision. pp. 6182--6191 (2019)

\bibitem{KYS}
Bhat, G., Danelljan, M., Gool, L.V., Timofte, R.: Know your surroundings:
  Exploiting scene information for object tracking. In: European Conference on
  Computer Vision. pp. 205--221. Springer (2020)

\bibitem{albumentations}
Buslaev, A., Iglovikov, V.I., Khvedchenya, E., Parinov, A., Druzhinin, M.,
  Kalinin, A.A.: Albumentations: fast and flexible image augmentations.
  Information  \textbf{11}(2), ~125 (2020)

\bibitem{TransT}
Chen, X., Yan, B., Zhu, J., Wang, D., Yang, X., Lu, H.: Transformer tracking.
  In: Proceedings of the IEEE/CVF Conference on Computer Vision and Pattern
  Recognition. pp. 8126--8135 (2021)

\bibitem{AFOD}
Chen, Y., Xu, J., Yu, J., Wang, Q., Yoo, B., Han, J.J.: Afod: Adaptive focused
  discriminative segmentation tracker. In: European Conference on Computer
  Vision. pp. 666--682. Springer (2020)

\bibitem{detnas}
Chen, Y., Yang, T., Zhang, X., Meng, G., Xiao, X., Sun, J.: Detnas: Backbone
  search for object detection. Advances in Neural Information Processing
  Systems  \textbf{32},  6642--6652 (2019)

\bibitem{coreml}
{Core ML}. \url{https://developer.apple.com/documentation/coreml}

\bibitem{ATOM}
Danelljan, M., Bhat, G., Khan, F.S., Felsberg, M.: Atom: Accurate tracking by
  overlap maximization. In: Proceedings of the IEEE/CVF Conference on Computer
  Vision and Pattern Recognition. pp. 4660--4669 (2019)

\bibitem{ECO}
Danelljan, M., Bhat, G., Shahbaz~Khan, F., Felsberg, M.: Eco: Efficient
  convolution operators for tracking. In: Proceedings of the IEEE conference on
  computer vision and pattern recognition. pp. 6638--6646 (2017)

\bibitem{prDIMP}
Danelljan, M., Gool, L.V., Timofte, R.: Probabilistic regression for visual
  tracking. In: Proceedings of the IEEE/CVF conference on computer vision and
  pattern recognition. pp. 7183--7192 (2020)

\bibitem{ImageNet}
Deng, J., Dong, W., Socher, R., Li, L.J., Li, K., Fei-Fei, L.: Imagenet: A
  large-scale hierarchical image database. In: 2009 IEEE conference on computer
  vision and pattern recognition. pp. 248--255. Ieee (2009)

\bibitem{ViT}
Dosovitskiy, A., Beyer, L., Kolesnikov, A., Weissenborn, D., Zhai, X.,
  Unterthiner, T., Dehghani, M., Minderer, M., Heigold, G., Gelly, S.,
  Uszkoreit, J., Houlsby, N.: An image is worth 16x16 words: Transformers for
  image recognition at scale. ICLR  (2021)

\bibitem{LaSOT}
Fan, H., Lin, L., Yang, F., Chu, P., Deng, G., Yu, S., Bai, H., Xu, Y., Liao,
  C., Ling, H.: Lasot: A high-quality benchmark for large-scale single object
  tracking. In: Proceedings of the IEEE/CVF Conference on Computer Vision and
  Pattern Recognition (CVPR) (June 2019)

\bibitem{WordNet}
Fellbaum, C.: WordNet: An Electronic Lexical Database. Bradford Books (1998)

\bibitem{autonomous}
Gao, M., Jin, L., Jiang, Y., Guo, B.: Manifold siamese network: A novel visual
  tracking convnet for autonomous vehicles. IEEE Transactions on Intelligent
  Transportation Systems  \textbf{21}(4),  1612--1623 (2020)

\bibitem{SqueezeNext}
Gholami, A., Kwon, K., Wu, B., Tai, Z., Yue, X., Jin, P., Zhao, S., Keutzer,
  K.: Squeezenext: Hardware-aware neural network design. In: Proceedings of the
  IEEE Conference on Computer Vision and Pattern Recognition (CVPR) Workshops
  (June 2018)

\bibitem{ResNet}
He, K., Zhang, X., Ren, S., Sun, J.: Deep residual learning for image
  recognition. In: Proceedings of the IEEE conference on computer vision and
  pattern recognition. pp. 770--778 (2016)

\bibitem{GOTURN}
Held, D., Thrun, S., Savarese, S.: Learning to track at 100 fps with deep
  regression networks. In: European conference on computer vision. pp.
  749--765. Springer (2016)

\bibitem{KCF}
Henriques, J.F., Caseiro, R., Martins, P., Batista, J.: High-speed tracking
  with kernelized correlation filters. IEEE transactions on pattern analysis
  and machine intelligence  \textbf{37}(3),  583--596 (2014)

\bibitem{TripletLoss}
Hoffer, E., Ailon, N.: Deep metric learning using triplet network. In:
  International workshop on similarity-based pattern recognition. pp. 84--92.
  Springer (2015)

\bibitem{MobileNet}
Howard, A.G., Zhu, M., Chen, B., Kalenichenko, D., Wang, W., Weyand, T.,
  Andreetto, M., Adam, H.: Mobilenets: Efficient convolutional neural networks
  for mobile vision applications. arXiv preprint arXiv:1704.04861  (2017)

\bibitem{GOT10k}
Huang, L., Zhao, X., Huang, K.: Got-10k: A large high-diversity benchmark for
  generic object tracking in the wild. IEEE Transactions on Pattern Analysis
  and Machine Intelligence  \textbf{43}(5),  1562--1577 (2021)

\bibitem{SqueezeNet}
Iandola, F.N., Han, S., Moskewicz, M.W., Ashraf, K., Dally, W.J., Keutzer, K.:
  Squeezenet: Alexnet-level accuracy with 50x fewer parameters and $<$0.5mb
  model size. arXiv:1602.07360  (2016)

\bibitem{BatchNorm}
Ioffe, S., Szegedy, C.: Batch normalization: Accelerating deep network training
  by reducing internal covariate shift. In: International conference on machine
  learning. pp. 448--456. PMLR (2015)

\bibitem{thermalstate}
{iOS thermal state}.
  \url{https://developer.apple.com/documentation/foundation/processinfo/thermalstate}

\bibitem{nfs}
Kiani~Galoogahi, H., Fagg, A., Huang, C., Ramanan, D., Lucey, S.: Need for
  speed: A benchmark for higher frame rate object tracking. In: Proceedings of
  the IEEE International Conference on Computer Vision. pp. 1125--1134 (2017)

\bibitem{ADAM}
Kingma, D.P., Ba, J.: Adam: A method for stochastic optimization. arXiv
  preprint arXiv:1412.6980  (2014)

\bibitem{Siam2015}
Koch, G., Zemel, R., Salakhutdinov, R., et~al.: Siamese neural networks for
  one-shot image recognition. In: ICML deep learning workshop. vol.~2, p.~0.
  Lille (2015)

\bibitem{VOT2020}
Kristan, M., Leonardis, A., Matas, J., Felsberg, M., Pflugfelder, R.,
  K{\"a}m{\"a}r{\"a}inen, J.K., Danelljan, M., Zajc, L.{\v{C}}.,
  Luke{\v{z}}i{\v{c}}, A., Drbohlav, O., et~al.: The eighth visual object
  tracking vot2020 challenge results. In: European Conference on Computer
  Vision. pp. 547--601. Springer (2020)

\bibitem{VOT}
Kristan, M., Matas, J., Leonardis, A., Vojir, T., Pflugfelder, R., Fernandez,
  G., Nebehay, G., Porikli, F., \v{C}ehovin, L.: A novel performance evaluation
  methodology for single-target trackers. IEEE Transactions on Pattern Analysis
  and Machine Intelligence  \textbf{38}(11),  2137--2155 (Nov 2016)

\bibitem{AvgPool}
Lee, C.Y., Gallagher, P.W., Tu, Z.: Generalizing pooling functions in
  convolutional neural networks: Mixed, gated, and tree. In: Proceedings of the
  19th International Conference on Artificial Intelligence and Statistics. pp.
  464--472 (2016)

\bibitem{SiamRPN++}
Li, B., Wu, W., Wang, Q., Zhang, F., Xing, J., Yan, J.: Siamrpn++: Evolution of
  siamese visual tracking with very deep networks. In: Proceedings of the
  IEEE/CVF Conference on Computer Vision and Pattern Recognition. pp.
  4282--4291 (2019)

\bibitem{SiamRPN}
Li, B., Yan, J., Wu, W., Zhu, Z., Hu, X.: High performance visual tracking with
  siamese region proposal network. In: Proceedings of the IEEE Conference on
  Computer Vision and Pattern Recognition (CVPR) (June 2018)

\bibitem{FocalLoss}
Lin, T.Y., Goyal, P., Girshick, R.B., He, K., Doll{\'a}r, P.: Focal loss for
  dense object detection. 2017 IEEE International Conference on Computer Vision
  (ICCV) pp. 2999--3007 (2017)

\bibitem{COCO}
Lin, T.Y., Maire, M., Belongie, S., Hays, J., Perona, P., Ramanan, D., Dollar,
  P., Zitnick, L.: Microsoft coco: Common objects in context. In: ECCV
  (September 2014)

\bibitem{RPT}
Ma, Z., Wang, L., Zhang, H., Lu, W., Yin, J.: Rpt: Learning point set
  representation for siamese visual tracking. In: European Conference on
  Computer Vision. pp. 653--665. Springer (2020)

\bibitem{survey2}
Marvasti-Zadeh, S.M., Cheng, L., Ghanei-Yakhdan, H., Kasaei, S.: Deep learning
  for visual tracking: A comprehensive survey. IEEE Transactions on Intelligent
  Transportation Systems  (2021)

\bibitem{trackingnet}
Muller, M., Bibi, A., Giancola, S., Alsubaihi, S., Ghanem, B.: Trackingnet: A
  large-scale dataset and benchmark for object tracking in the wild. In:
  Proceedings of the European Conference on Computer Vision (ECCV). pp.
  300--317 (2018)

\bibitem{MDNet}
Nam, H., Han, B.: Learning multi-domain convolutional neural networks for
  visual tracking. In: Proceedings of the IEEE conference on computer vision
  and pattern recognition. pp. 4293--4302 (2016)

\bibitem{pytorch}
Paszke, A., Gross, S., Massa, F., Lerer, A., Bradbury, J., Chanan, G., Killeen,
  T., Lin, Z., Gimelshein, N., Antiga, L., et~al.: Pytorch: An imperative
  style, high-performance deep learning library. Advances in neural information
  processing systems  \textbf{32} (2019)

\bibitem{nas}
Pham, H., Guan, M., Zoph, B., Le, Q., Dean, J.: Efficient neural architecture
  search via parameters sharing. In: International Conference on Machine
  Learning. pp. 4095--4104. PMLR (2018)

\bibitem{YouTubeBB}
Real, E., Shlens, J., Mazzocchi, S., Pan, X., Vanhoucke, V.:
  Youtube-boundingboxes: A large high-precision human-annotated data set for
  object detection in video. In: proceedings of the IEEE Conference on Computer
  Vision and Pattern Recognition. pp. 5296--5305 (2017)

\bibitem{IoUloss}
Rezatofighi, H., Tsoi, N., Gwak, J., Sadeghian, A., Reid, I., Savarese, S.:
  Generalized intersection over union. In: The IEEE Conference on Computer
  Vision and Pattern Recognition (CVPR) (June 2019)

\bibitem{robotics}
Robin, C., Lacroix, S.: Multi-robot target detection and tracking: taxonomy and
  survey. Autonomous Robots  \textbf{40}(4),  729--760 (2016)

\bibitem{mobilenetv2}
Sandler, M., Howard, A., Zhu, M., Zhmoginov, A., Chen, L.C.: Mobilenetv2:
  Inverted residuals and linear bottlenecks. In: Proceedings of the IEEE
  conference on computer vision and pattern recognition. pp. 4510--4520 (2018)

\bibitem{WAP}
Shin, H., Cho, H., Kim, D., Ko, D.K., Lim, S.C., Hwang, W.: Sequential
  image-based attention network for inferring force estimation without haptic
  sensor. IEEE Access  \textbf{7},  150237--150246 (2019)

\bibitem{SINT}
Tao, R., Gavves, E., Smeulders, A.W.: Siamese instance search for tracking. In:
  Proceedings of the IEEE conference on computer vision and pattern
  recognition. pp. 1420--1429 (2016)

\bibitem{FCOS}
Tian, Z., Shen, C., Chen, H., He, T.: Fcos: Fully convolutional one-stage
  object detection. In: Proceedings of the IEEE/CVF international conference on
  computer vision. pp. 9627--9636 (2019)

\bibitem{Matas}
Voj{\'i}r, T., Noskova, J., Matas, J.: Robust scale-adaptive mean-shift for
  tracking. In: SCIA (2013)

\bibitem{TrMeetsTr}
Wang, N., Zhou, W., Wang, J., Li, H.: Transformer meets tracker: Exploiting
  temporal context for robust visual tracking. In: Proceedings of the IEEE/CVF
  Conference on Computer Vision and Pattern Recognition. pp. 1571--1580 (2021)

\bibitem{RASNet}
Wang, Q., Teng, Z., Xing, J., Gao, J., Hu, W., Maybank, S.: Learning
  attentions: residual attentional siamese network for high performance online
  visual tracking. In: Proceedings of the IEEE conference on computer vision
  and pattern recognition. pp. 4854--4863 (2018)

\bibitem{FBNet}
Wu, B., Dai, X., Zhang, P., Wang, Y., Sun, F., Wu, Y., Tian, Y., Vajda, P.,
  Jia, Y., Keutzer, K.: Fbnet: Hardware-aware efficient convnet design via
  differentiable neural architecture search. The IEEE Conference on Computer
  Vision and Pattern Recognition (CVPR)  (2019)

\bibitem{shift}
Wu, B., Wan, A., Yue, X., Jin, P., Zhao, S., Golmant, N., Gholaminejad, A.,
  Gonzalez, J., Keutzer, K.: Shift: A zero flop, zero parameter alternative to
  spatial convolutions. In: Proceedings of the IEEE Conference on Computer
  Vision and Pattern Recognition. pp. 9127--9135 (2018)

\bibitem{OTB}
Wu, Y., Lim, J., Yang, M.H.: Online object tracking: A benchmark. In:
  Proceedings of the IEEE Conference on Computer Vision and Pattern Recognition
  (CVPR) (June 2013)

\bibitem{surveillance}
Xing, J., Ai, H., Lao, S.: Multiple human tracking based on multi-view
  upper-body detection and discriminative learning. In: 2010 20th International
  Conference on Pattern Recognition. pp. 1698--1701. IEEE (2010)

\bibitem{xu2021regnet}
Xu, J., Pan, Y., Pan, X., Hoi, S., Yi, Z., Xu, Z.: Regnet: self-regulated
  network for image classification. IEEE Transactions on Neural Networks and
  Learning Systems  (2022)

\bibitem{LADCF}
Xu, T., Feng, Z.H., Wu, X.J., Kittler, J.: Learning adaptive discriminative
  correlation filters via temporal consistency preserving spatial feature
  selection for robust visual object tracking. IEEE Transactions on Image
  Processing  \textbf{28}(11),  5596--5609 (2019)

\bibitem{SiamFC++}
Xu, Y., Wang, Z., Li, Z., Yuan, Y., Yu, G.: Siamfc++: Towards robust and
  accurate visual tracking with target estimation guidelines. In: Proceedings
  of the AAAI Conference on Artificial Intelligence. vol.~34, pp. 12549--12556
  (2020)

\bibitem{STARK}
Yan, B., Peng, H., Fu, J., Wang, D., Lu, H.: Learning spatio-temporal
  transformer for visual tracking. arXiv preprint arXiv:2103.17154  (2021)

\bibitem{stark_lightning}
Yan, B., Peng, H., Fu, J., Wang, D., Lu, H.: Stark lightning.
  \url{https://github.com/researchmm/Stark} (2021)

\bibitem{LightTrack}
Yan, B., Peng, H., Wu, K., Wang, D., Fu, J., Lu, H.: Lighttrack: Finding
  lightweight neural networks for object tracking via one-shot architecture
  search. In: CVPR 2021 (June 2021)

\bibitem{AlphaRef}
Yan, B., Zhang, X., Wang, D., Lu, H., Yang, X.: Alpha-refine: Boosting tracking
  performance by precise bounding box estimation. In: Proceedings of the
  IEEE/CVF Conference on Computer Vision and Pattern Recognition. pp.
  5289--5298 (2021)

\bibitem{AR}
Zhang, G., Vela, P.A.: Good features to track for visual slam. In: Proceedings
  of the IEEE conference on computer vision and pattern recognition. pp.
  1373--1382 (2015)

\bibitem{shufflenet}
Zhang, X., Zhou, X., Lin, M., Sun, J.: Shufflenet: An extremely efficient
  convolutional neural network for mobile devices. In: Proceedings of the IEEE
  conference on computer vision and pattern recognition. pp. 6848--6856 (2018)

\bibitem{OceanPlus}
Zhang, Z., Li, B., Hu, W., Peng, H.: Towards accurate pixel-wise object
  tracking by attention retrieval. arXiv preprint arXiv:2008.02745  (2020)

\bibitem{SiamDW}
Zhang, Z., Peng, H.: Deeper and wider siamese networks for real-time visual
  tracking. In: Proceedings of the IEEE/CVF Conference on Computer Vision and
  Pattern Recognition. pp. 4591--4600 (2019)

\bibitem{Ocean}
Zhang, Z., Peng, H., Fu, J., Li, B., Hu, W.: Ocean: Object-aware anchor-free
  tracking. In: Computer Vision--ECCV 2020: 16th European Conference, Glasgow,
  UK, August 23--28, 2020, Proceedings, Part XXI 16. pp. 771--787. Springer
  (2020)

\bibitem{DCFST}
Zheng, L., Tang, M., Chen, Y., Wang, J., Lu, H.: Learning feature embeddings
  for discriminant model based tracking. In: European Conference on Computer
  Vision. pp. 759--775. Springer (2020)

\bibitem{DaSiamRPN}
Zhu, Z., Wang, Q., Li, B., Wu, W., Yan, J., Hu, W.: Distractor-aware siamese
  networks for visual object tracking. In: Proceedings of the European
  Conference on Computer Vision (ECCV). pp. 101--117 (2018)

\end{thebibliography}

\newpage
\section*{Appendix A. Training Datasets}

The \textbf{YouTube-BoundingBoxes} \cite{YouTubeBB} is a large-scale dataset of videos.
The dataset consists of approximately \textbf{380,000} video segments of 15-20s
with a recording quality often akin to that of a hand-held cell phone camera.

The \textbf{LaSOT} \cite{LaSOT} 
consists of 1,400 sequences with more than \textbf{3.5M} frames in total. 
Each sequence contains 2,500 frames on average and the dataset represents 70 different object categories.

The \textbf{GOT-10k} \cite{GOT10k} is built upon the backbone of WordNet structure \cite{WordNet} and it populates the majority of over 560 classes of moving objects and 87 motion patterns. It contains more than 10,000 of short video sequences with  more than \textbf{1.5M} manually labeled bounding boxes, annotated at 30 frames per second, enabling unified training and stable evaluation of deep trackers. 


The \textbf{ImageNet-VID} \cite{ImageNet} is a benchmark created for video object detection task. It contains 30 object categories. 
Overall, benchmark consists of near \textbf{2M} annotations and over 4,000 video sequences.

In addition, similar to other tracking models \cite{SiamFC}, \cite{DaSiamRPN}, \cite{Ocean}, we use a part of the \textbf{COCO} \cite{COCO} dataset for object detection with 80 different object categories to diversify the training dataset for visual object tracking. In our setup, we set $I_S = I_T$ to let the network efficiently predict the object's location in a larger context.

\section*{Appendix B. Technical details}
\subsection*{B.1. Pixel-wise correlation implementation}
Classical cross-correlation cannot be executed by most mobile neural network inference engines such as CoreML \cite{coreml} due to unsupported convolutional operation with dynamic weights from the template features. 
Thus, we reformulated the pixel-wise cross-correlation operation as a matrix multiplication operation that is better supported on mobile devices. 

Given input image features $\Phi_{S}$ and template image features $\Phi_{T}$ flattened along the spatial dimensions to shapes $C \times WH$ and $C \times wh$ respectively, we compute pixel-wise cross-correlation features $\Phi_{corr}$ as:
\begin{equation}
\Phi_{corr} = \Phi_{T}^{\top} \Phi_{S}
\end{equation}
The resulting $\Phi_{corr}$ will be a tensor of shape $wh \times WH$.

\subsection*{B.2. Smartphone-based Implementation} The models are trained offline using PyTorch \cite{pytorch} and then ported with an optimal model snapshot to mobile devices for inference. All models are executed in \textit{float16} mode for faster execution comparing to \textit{float32} computations. The precision loss of \textit{float16} computations is negligible, we observe that the results differ only by $\pm0.5\%$ depending on the experiment. 

We use Core ML \cite{coreml} framework to run FEAR tracker on iPhone devices. Core ML is a machine learning API from Apple that optimizes on-device neural network inference by leveraging the CPU, GPU and Neural Engine. 

For Android devices, we employ TensorFlow Lite \cite{tensorflow2015-whitepaper} which is an open-source deep learning framework for on-device inference from Google supporting execution on CPU, GPU and DSP.

\section*{Appendix C. Qualitative comparison}
\begin{figure*}[htp!]
\centering
  \includegraphics[width=0.99\textwidth]{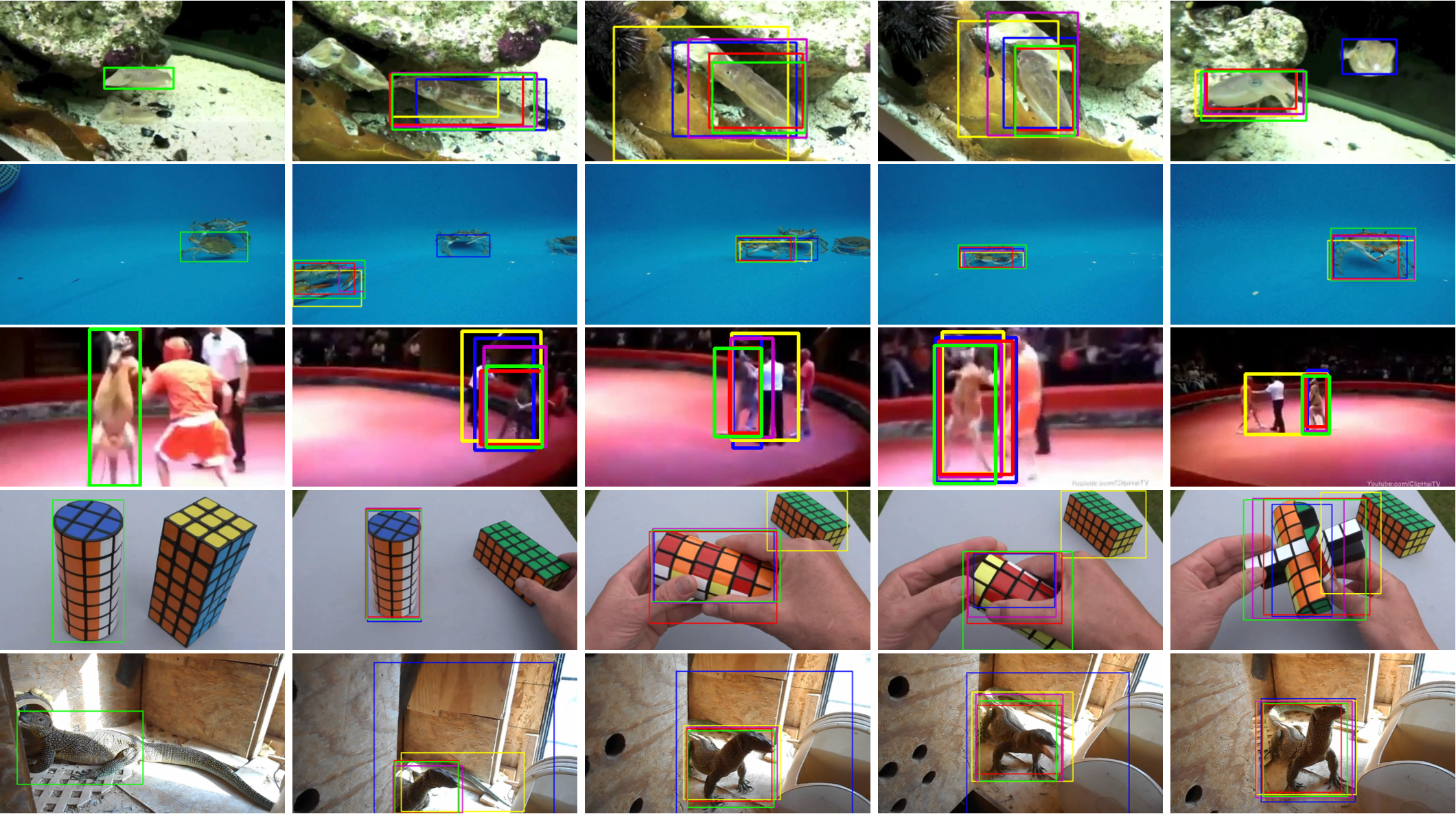} \\
   \caption{Qualitative comparison of FEAR tracker with state-of-the-art methods on challenging cases of variations in tracked object appearance from LaSOT benchmark \cite{LaSOT}. \textcolor{green}{Green}: Ground Truth, \textcolor{red}{Red}: FEAR-L, \textcolor{yellow}{Yellow}: STARK Lightning, \textcolor{blue}{Blue}: Ocean, \textcolor{purple}{Purple}: Stark-ST50.}
   \label{fig:comparison}
\end{figure*}

The comparison of FEAR tracker with the state-of-the-art methods is presented in Figure \ref{fig:comparison}. We display the tracking results of every 200 frames (0 - 1000) on the challenging cases from LaSOT benchmark where the object appearance and scale change throughout the video. 

\end{document}